\documentclass[10pt,conference,a4paper]{IEEEtran}

\ifCLASSINFOpdf
\usepackage[pdftex]{graphicx}
\else
\fi

\usepackage{amssymb}
\usepackage{amsmath}
\usepackage{latexsym}   

\usepackage{booktabs}
\usepackage{stfloats}
\usepackage{cite}
\usepackage{color}
\usepackage{adjustbox}
\usepackage{multirow}
\usepackage{blindtext}
\usepackage{graphicx}
\usepackage{caption}
\begin{document}

\title{Unsupervised Contrastive Photo-to-Caricature Translation based on Auto-distortion\\ }

\author{\IEEEauthorblockN{Yuhe Ding\IEEEauthorrefmark{1}\IEEEauthorrefmark{2},
		Xin Ma\IEEEauthorrefmark{2}\IEEEauthorrefmark{3},
		Mandi Luo\IEEEauthorrefmark{2}\IEEEauthorrefmark{3},
		Aihua Zheng\IEEEauthorrefmark{1} and
		Ran He\IEEEauthorrefmark{2}\IEEEauthorrefmark{3}}
	
	\IEEEauthorblockA{\IEEEauthorrefmark{1}School of Computer Science and Technology, Anhui University}
	\IEEEauthorblockA{\IEEEauthorrefmark{2}NLPR$\;\&\;$CEBSIT, CASIA}
	\IEEEauthorblockA{\IEEEauthorrefmark{3}School of Artificial Intelligence, University of Chinese Academy of Sciences  }
	
	\IEEEauthorblockA{
		Email: madao3c@foxmail.com, xin.ma@cripac.ia.ac.cn, mandi.luo@cripac.ia.ac.cn, ahzheng@foxmail.com, rhe@nlpr.ia.ac.cn
	}
}




\maketitle
 \footnotetext[1]{Yuhe Ding and Xin Ma contribute equally to this work.}
\begin{abstract}
	Photo-to-caricature translation aims to synthesize the caricature as a rendered image exaggerating the features through sketching, pencil strokes, or other artistic drawings. 
	Style rendering and geometry deformation are the most important aspects in photo-to-caricature translation task.
	To take both into consideration, we propose an unsupervised contrastive photo-to-caricature translation architecture. 
	Considering the intuitive artifacts in the existing methods, we propose a contrastive style loss for style rendering to enforce the similarity between the style of rendered photo and the caricature, and simultaneously enhance its discrepancy to the photos.
	To obtain an exaggerating deformation in an unpaired/unsupervised fashion, we propose a Distortion Prediction Module (DPM) to predict a set of displacements vectors for each input image while fixing some controlling points, followed by the thin plate spline interpolation for warping.
	The model is trained on unpaired photo and caricature while can offer bidirectional synthesizing via inputting either a photo or a caricature.
	Extensive experiments demonstrate that the proposed model is effective to generate hand-drawn like caricatures compared with existing competitors. 
	
\end{abstract}


%
\IEEEpeerreviewmaketitle

  \section{Introduction}\label{intro}
%
As a special image-to-image translation task, caricature generation requires exaggerating on face features, and re-rendering the facial texture to form a portrait.
Existing methods mainly fall into three classes, deformation-based, texture-based, and methods taking both aspects into consideration.

Deformation-based methods focus on the geometric distortion by using a certain guidance, such as 2D landmarks or 3D mesh~\cite{han2018caricatureshop}.
However, it is challenging to guarantee the precise guidances.
Furthermore, without consideration of texture render, they tend to generate less hand-drawn painting like results.

Texture-based methods devote to obtain caricature's style via the prevalent GANs~\cite{goodfellow2014generative}.
Zheng~\textit{et al}.~\cite{zheng2019unpaired} use the cycle generators to preserve the texture consistency in caricature generation. 
Benefit from facial masks, Li~\textit{et al}.~\cite{li2018carigan} transfer the texture of the input image through weakly paired adversarial learning.
However, they only consider the geometric deformation in representation space thus result in limited deformation. 

More recently, to exaggerate deformation and obtain plausible texture simultaneously,
Cao~\textit{et al}.~\cite{cao2018carigans} propose to first employ two CycleGANs~\cite{zhu2017unpaired} to transfer photos and landmarks, respectively, then warp the rendered image with the help of landmarks. 
Shi~\textit{et al}.~\cite{shi2019warpgan} propose a warp controller to predict a set of controlling points and its displacements,
followed by an AdaIN-based~\cite{huang2017arbitrary} rendering network to transfer the texture. 
Despite the great progress for caricature translation by taking both deformation and texture into consideration in these two methods, there are two intuitive defects remaining improvement. 

On the one hand, their style rendering effect appears either not plausible enough or with prominent artifacts.
Cao~\textit{et al}.~\cite{cao2018carigans} retain too much photo information since the CycleGAN-based architecture emphasizes single style translation learning, which is hard to learn the diverse texture styles existing in caricatures. 
Shi~\textit{et al}.~\cite{shi2019warpgan} tend to generate more artifacts since
the AdaIN-based structure assumes that feature maps in different channels are uncorrelated, which ignores the global information in style rendering.
Therefore, we propose a contrastive photo-caricature translation method in this paper.
First, we design our transferring architecture based on weight sharing strategy~\cite{liu2017unsupervised} to maintain the global information without the uncorrelated assumption between different channels.
Second, Hadsell~\textit{et al}.~\cite{hadsell2006dimensionality} evidences that contrastive loss can pull closer similar pairs and push away dissimilar pairs,
 However, conventional Euclidean distance based contrastive loss is simply defined by subtraction between two images, which is not suitable to measure the similarity of image styles.
Therefore, we propose 
 a style distance by gram matrix, which can enhance the texture details by calculating the dot product of feature maps for any two channels,
and introduce the proposed contrastive style loss to enforce the texture similarity of the rendered photo to caricatures and its discrepancy to the photos.

On the other hand, 
%
although Cao~\textit{et al}.~\cite{cao2018carigans} support unsupervised translation due to their cycle architecture, the precise landmarks required as guidance during the translation are hard to be guaranteed.
Shi~\textit{et al}.~\cite{shi2019warpgan} propose a warp controller to predict a set of points for warping, which avoids extra guidance information such as landmarks. 
However, as the weakly supervision information, the identity labels are hard to obtain in the wild, which is the crucial information in Shi~\textit{et al}.~\cite{shi2019warpgan} to generate characteristic exaggeration. 
To obtain exaggerating deformation in the unsupervised fashion without the guidance condition, we propose the Distortion Prediction Module (DPM), to automatically predict a set of displacements for the predefined controlling points in this paper. 
Then we use the controlling points and their displacements for warping via a classical interpolation method: thin plate spline~\cite{bookstein1989principal}. 
Note that since we only predict the displacements rather than simultaneously predict the controlling points, we can decrease the unexpected deformation as well as simplify our network complexity.
Particularly, in the test stage, random-perturbed photos are input to DPM to achieve diverse deformation in caricatures.
Note that our goal is the unsupervised/unpaired translation which means identity label is not available/required.
Furthermore, it is available for bidirectional synthesizing with either a photo or a caricature as the input.
More comparisons to the state-of-the-art caricature generation methods are shown in Table~\ref{table1}.

Based on the above discussion, we propose an unsupervised Contrastive Translation for auto-distortion Photo-to-Caricature translation method in this paper. %
The main contributions include:
\begin{itemize}
	\item To reduce artifacts in rendered photos, we propose a novel contrastive loss by define a style to enforce the similarity between the rendered photo’s texture and caricatures, and enforce its discrepancy to the photos, which can generate plausible textures in the rendered photos with more details.
	\item To obtain exaggerating deformation in the unpaired setting, we propose a new symmetrical architecture with Distortion Prediction Modules (DPM), which predicts a set of displacements vectors without any guidance to warp the images in an unsupervised fashion. 
	\item Experiments on the benchmark caricature generation dataset WebCaricature\cite{huo2017webcaricature} compared to the state-of-the-art methods demonstrate our methods can synthesize caricatures with more hand-drawn like texture with diverse deformation without guidance in the unsupervised fashion. In addition, our method supports bidirectional translation due to the symmetrical architecture.
\end{itemize}

\begin{figure*}[h]
	\centering
	\includegraphics[scale=0.54]{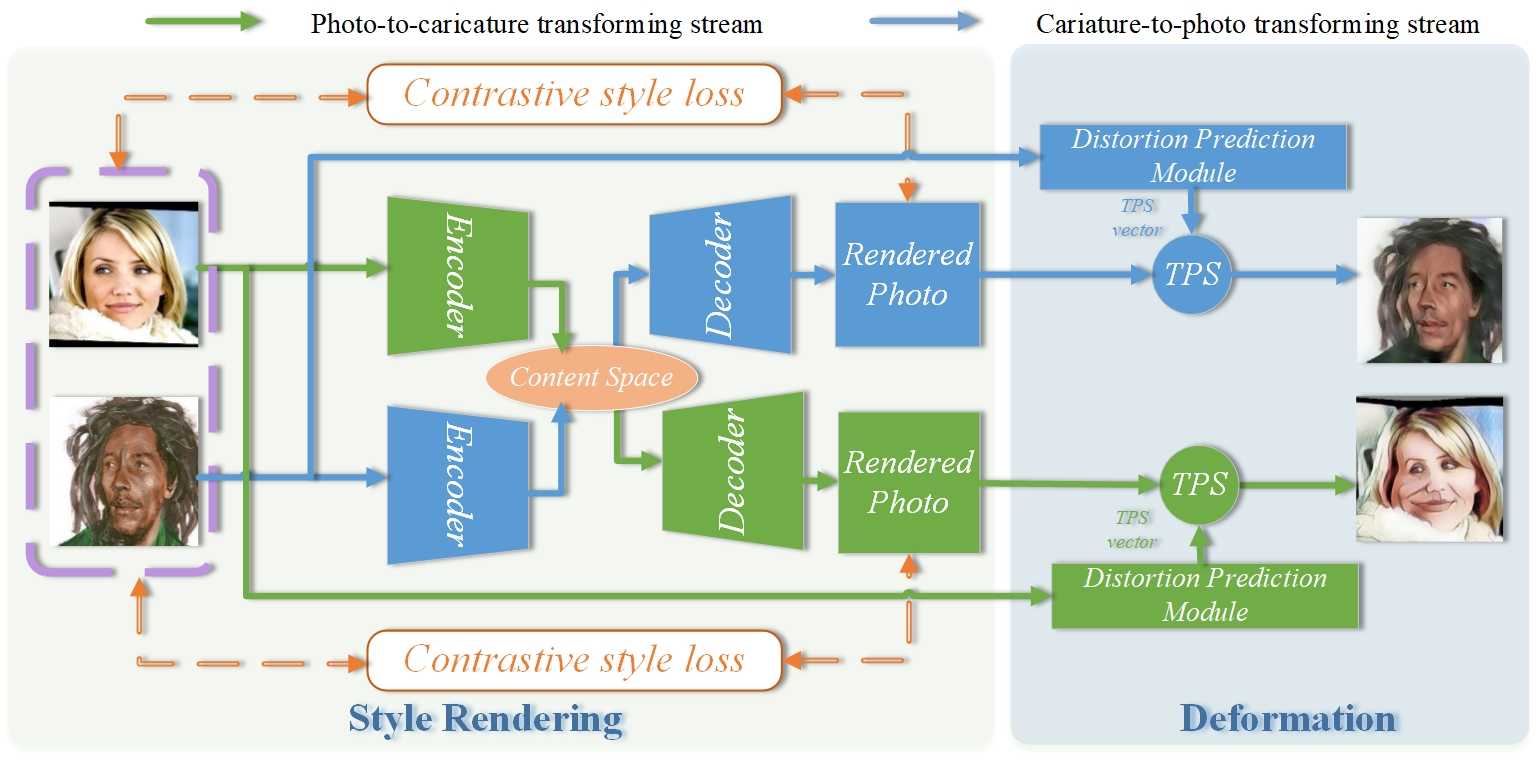} 
	\captionof{figure}{Overview of our symmetric architecture. The green and blue lines represent photo-to-caricature and caricature-to-photo transforming streams respectively. $Content$ $Space$ represents the common content space. Our translation fall into two stages: style rendering and deformation. We propose contrastive style loss and distortion prediction module in these two stages, respectively.}\label{net}
\end{figure*}

\section{Related works}\label{relatedworks}
We briefly review the related works on the following three aspects. 

\subsection{Caricature Generation}
To generate exaggerating caricatures, traditional works define a shape representation such as 2D landmarks and 3D mesh~\cite{lewiner2011interactive, han2018caricatureshop}, then calculate a mean face to exaggerate the representative feature where have the largest deviation from mean face.
However, their capability of geometry distortion are generally limited due to the need for guidance. And their results suffer from poor visual quality because these networks are not suitable for problems with large spatial variation.
With the blossom of deep learning, generative adversarial networks (GAN)~\cite{goodfellow2014generative} have a widespread application in computer vision~\cite{zhang2017demeshnet,he2009robust}, especially in caricature generation.
%
Cao \textit{et al}.~\cite{cao2018carigans} recently propose to decouple texture rendering and geometric deformation with two CycleGANs trained on image and landmark space, respectively.
But with their face shape modeled in the PCA subspace of landmarks, they suffer from the same problem of the traditional deformation-based methods
Shi \textit{et al}.~\cite{shi2019warpgan} propose a weakly-supervised generation model, and obtain commendable results in the aspect of geometry distortion. A warp controller and a rendering network are used to process geometry and style, respectively. 
The style rendering network is based on AdaIN~\cite{huang2017arbitrary}, which assumes the feature maps in different channels are uncorrelated, hence ignoring the global and identical information. Therefore, it tends to produce many artifacts.

\subsection{Style Rendering}
Many style translation tasks have been proposed to render images from the source texture to the target one.
These methods fall into two main categories: supervised and unsupervised models. The major difference of the two categories is whether the training data are paired or not.
Pix2pix~\cite{isola2017image} is one of the prevalent supervised frameworks, which learns a mapping function from the source to the target domain.
Wang \textit{et al}. further propose Pix2pixHD~\cite{wangpix2pixhd} for high-resolution photo-realism translation.  
Representative supervised models ~\cite{isola2017image,wangpix2pixhd,lu2018image,zhu2017toward} are wildly researched and applied in the past decade.
However, paired image data restrict its application in real-world applications.
As pioneer unsupervised translation, CycleGAN~\cite{zhu2017unpaired}, DualGAN~\cite{yi2017dualgan} and DiscoGAN~\cite{kim2017learning} translate images using cycle consistency.
Subsequently, various GAN-based unsupervised translation models~\cite{choi2018stargan, benaim2017one, anoosheh2018combogan, tang2018dual, wang2018mix} emerge for one-to-many translation, many-to-many translation and so on.
However, they mainly focus on style rendering and less consider the spatial distortion that is important for photo-to-caricature translation.
\begin{table}[hb]
	\renewcommand\arraystretch{1.1}
	\renewcommand\tabcolsep{0.5mm}
	\centering
	\resizebox{88mm}{19mm}{
		\begin{tabular}{lccccc} 
			\toprule
			\multirow{2}{*}{Method} & \multicolumn{5}{c}{Component} \\ \cline{2-6} 
			& Texture & Exaggeration & Diversity & Bidirectional & Unsupervised \\
			\midrule
			CycleGAN\cite{zhu2017unpaired}&  \checkmark & $\times$  & $\times$ & \checkmark  & \checkmark \\
			UNIT\cite{liu2017unsupervised} & \checkmark & $\times$  & $\times$ & \checkmark  & \checkmark \\
			MUNIT\cite{huang2018multimodal}& \checkmark & $\times$  & \checkmark & \checkmark  & \checkmark \\
			StarGAN\cite{choi2018stargan} & \checkmark & $\times$  & $\times$ & \checkmark  & \checkmark \\
			CariGAN\cite{li2018carigan} & \checkmark& \checkmark & $\times$& $\times$& $\times$\\
			CariGANs\cite{cao2018carigans} & \checkmark & \checkmark & $\times$ & \checkmark & \checkmark \\
			WarpGAN\cite{shi2019warpgan} & \checkmark & \checkmark & \checkmark & $\times$ &$\times$ \\
			Ours &\checkmark & \checkmark & \checkmark & \checkmark &\checkmark \\
			\bottomrule  
	\end{tabular}}
	\caption{Comparison of different caricature generation methods.} \label{table1}
\end{table}
\subsection{Spatial Distortion}
To process geometry distortion, parameter-based methods~\cite{jaderberg2015spatial,lin2018st} predict a set of deformation parameters, but they can not process with fine-grained distortion because the number of parameters is limited.
Some methods~\cite{ganin2016deepwarp,wu2019attribute} use a dense motion field to warp the images, all vertices in a deformation grid are predicted, yet most of them are useless.
Besides, the disentanglement-based method~\cite{detlefsen2019explicit} performs well at some simple datasets. However, because of the hundreds of styles, caricatures are too complex to disentangle.
To caricature's deformation, there are some specific approaches.
Cao \textit{et al}.~\cite{cao2018carigans} use PCA landmarks in a way of CycleGAN~\cite{zhu2017unpaired}, yet the manual annotations are needed.
Shi \textit{et al}.~\cite{shi2019warpgan} propose a warp controller to predict a set of control points and its displacements automatically. This method works well under a weakly supervised setting so that, it is not suitable for the unsupervised setting.

\begin{center}
	\centering
	\includegraphics[scale=0.45]{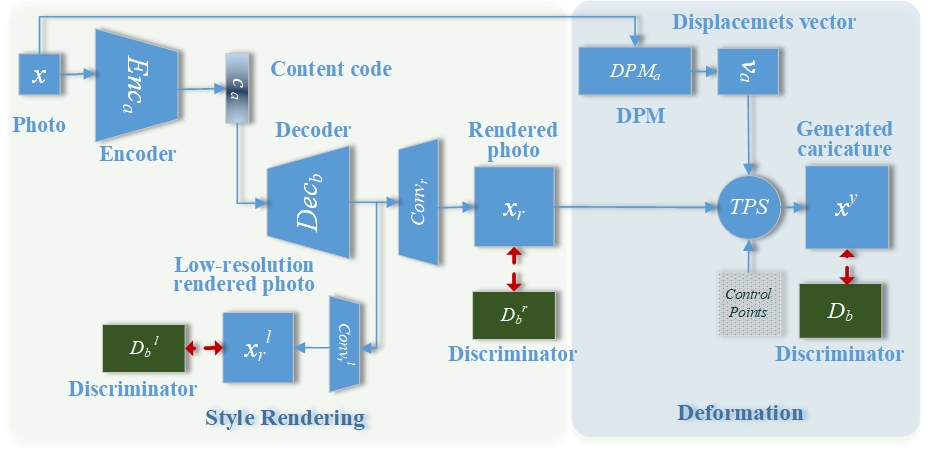} 
	\captionof{figure}{The pipeline of photo-to-caricature transforming. The caricature-to-photo transforming can be designed in the same manner. }\label{half}
\end{center}
\section{Method}\label{method}
In this paper, we propose a symmetrical encoder-decoder architecture with contrastive style loss and Distortion Prediction Module (DPM) for photo to caricature transforming, 
to decrease the rendered photo's artifacts,  
and exaggerating geometric deformation under unpaired setting.
%
%
%
%
Fig.~\ref{net} illustrates the whole architecture of the proposed model.  
It is a symmetrical structure that supports the bidirectional transforming. Each direction of transforming stream consists of an encoder, a decoder, a distortion prediction module (DPM) and three discriminators.

\subsection{Transforming Stream}\label{transStream}
Let $x\in X$ be the images from human face photos, and $y\in Y$ be the images from hand-drawn caricatures. 
Given an input photo $x$, our goal is to generate a corresponding caricature $x^y\in Y$. 
As shown in Fig.~\ref{half}, our transforming stream falls into two stages: style rendering and geometry deformation.

First, for an input face photo $x$, the encoder $Enc_{a}$ first maps $x$ to a code in content space, and then decodes the content code $c_a$ to transform the photo into rendered photo $x^r$ via decoder $Dec_b$. 
Note that an auxiliary rendered photo $x_r^l$ with a quarter size of rendered photo $x_r$ is output by different layers of decoder $Dec_b$ too.
To enforce the similarity between rendered photo and input caricature and its discrepancy to input photo, we propose a contrastive style loss in this stage.

Second, the distortion prediction module $DPM_{a}$ estimates the displacements vector $v_{a}$ for the input photo $x$,.
Given the predefined controlling points $p_0$, the deformed photo $x^y$ is obtained via thin plate spline interpolation (TPS)~\cite{bookstein1989principal},
\begin{equation} \label{x^y}
x^y = TPS_{p_{0},v_a}(Dec_{b}(c_{a})),
\end{equation}
where $v_{a}=DPM_{a}(x)$, $c_a\sim q_a(c_a|x)$, and Gaussian distribution $q_a(c_a|x)\equiv N(c_a|Enc_a(x),I)$. 

Note that we obtain the exaggerating deformation in the unpaired/unsupervised fashion without any additional guidance.

\subsection{Style Rendering}\label{stage 1}
Different with AdaIN~\cite{huang2017arbitrary} based methods~\cite{huang2018multimodal,shi2019warpgan}, which assume that different channel is uncorrelated, we use shared-latent space assumption and weight sharing strategy, to better capture the global information in photos.
Then, we propose to enforce a contrastive style loss to to decrease the artifacts.

\noindent
\textbf{Shared-latent space assumption.}
As noted in Liu \textit{et al.}~\cite{liu2017unsupervised}, for any given images $x$ and $y$ from different domains, there exists a shared latent code $z$ in a shared-latent space, as shown in Fig~\ref{content}.
Specifically, given encoders $Enc_a, Enc_b$, decoders $Dec_a, Dec_b$, we have $z=Enc_a(x)=Enc_b(y)$, and $x=Dec_a(z),y=Dec_b(z)$. 
With this assumption, we have $y=Dec_b(Enc_a(x))$, $x=Dec_a(Enc_b(y))$, 
which map $X$ to $Y$ and $Y$ to $X$, respectively. 

\begin{center}
	\centering
	\includegraphics[scale=0.5]{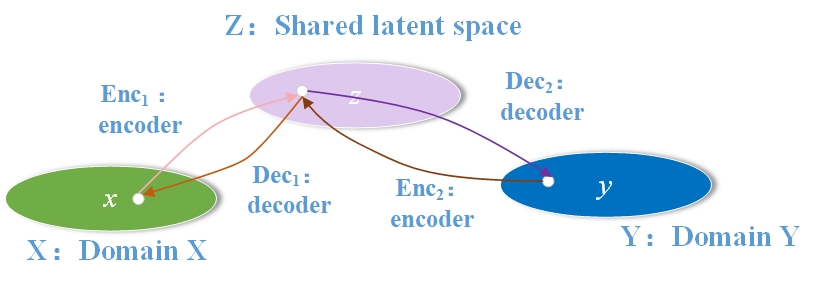} 
	\captionof{figure}{The shared-latent space assumption.}\label{content}
\end{center}

Our backbone is based on variational auto-encoders (VAEs)~\cite{kingma2013auto,rezende2014stochastic,larsen2015autoencoding} and GANs~\cite{goodfellow2014generative}~\cite{liu2017unsupervised}. 
For the photo domain $X$, encoder-decoder pair \{$Enc_a, Dec_a$\} constitute a VAE, while decoder-discriminator pairs \{$Dec_a, D_a^{l}$\}, \{$Dec_a,D_a^{r}$\}, and \{$Dec_a,D_a$\} constitute three GANs.
With the shared-latent space assumption, we assume that the content space $C$ is conditionally independent and Gaussian with unit variance.
%
%
The encoder outputs a mean vector $Enc_a(x)$ and the distribution of the content code $c_a$ is given by $q_a(c_a|x)\equiv N(c_a|Enc_a(x),I)$, where $I$ is an identity matrix. 

The reconstructed photo $\widetilde x$ is calculated by $\widetilde x=Dec_a(c_a\sim q_a(c_a|x))$, and rendered photo  $x_r$ is calculated by $x_r = Dec_b(c_a\sim q_a(c_a|x))$. Here the distribution of $q_a(c_a|x)$ is treated as a random vector sampled from $N(c_a|Enc_a(x),I)$.

The reparameterization trick~\cite{kingma2013auto} is utilized to reparameterize the non-differentiable sampling operation as a differentiable operation using auxiliary random variables. 
This reparameterization trick allows us to train VAEs using back-prop.
Let $s$ be a random vector with a multi-variate Gaussian distribution: $s\sim N(s|0,I)$. The sampling operations of content code $c_a\sim q_a(c_a|x)$ can be implemented via $c_a=Enc_a(x)+s$.
Similarly, reconstructed caricature $\widetilde y=Dec_b(c_b\sim q_b(c_b|y))$, caricature content code $c_b=Enc_b(y)+s$.

We now introduce the reconstructed loss within domain,

\begin{equation} \label{recon_loss}
\begin{aligned}
\mathcal{L}_{rec}=&{\parallel \widetilde x-x\parallel}_1 +{\parallel \widetilde y-y\parallel}_1 \\
= &{\parallel Dec_a(c_a)-x\parallel}_1+{\parallel Dec_b(c_b)-y\parallel}_1.
\end{aligned}
\end{equation}

Based on the shared-latent space assumption, we take a weight-sharing approach~\cite{liu2017unsupervised}. 
%
In detail, the last two layers' weights of encoder $Enc_a$ and encoder $Enc_b$, which extract high-level features, are sharing.
Analogously for the first two layers of decoder $Dec_a$ and decoder $Dec_b$ which are vital to decode the high-level representation. 
The KL loss is introduced to enforce the outputs of these two encoders share a common content space,
\begin{equation}\label{KL_loss} 
\begin{aligned}
\mathcal{L}_{KL}=&KL(q_a(c_a|x)\parallel p_s(c))+KL(q_b(c_b|y)\parallel p_s(c)),
\end{aligned}
\end{equation}
where {KL divergence term is defined as:}
\begin{equation}
KL(p\parallel q)= \sum_{i=1}^{n}p(x_i)\cdot log\frac{p(x_i)}{q(x_i)}  ,
\end{equation}

the KL divergence term penalizes the deviation of the distribution of the content space from the prior distribution. We use Gaussian distribution to model $q_a$ and $q_b$, where the prior distribution is a standard Gaussian distribution $p_s(c)\sim N(0,I)$.

In the photo-to-caricature translation stream, decoder $Dec_b$ decodes the content code of domain $X$. 
Note that our decoders have two branches, which generate images with 128-scale and 256-scale, noted as $Dec_a^{l}$ and $Dec_a^{r}$ respectively. 
%
%
To generate plausible rendered photo, the adversarial loss is introduced,
%
\begin{equation}\label{advG}
\begin{aligned}
\mathcal{L}_{adv}^{styG}=&\mathbb E[log(1-D_b^{l}(x_r^l))]+\mathbb E[log(1-D_b^{r}(x_r))]+\\
&\mathbb E[log(1-D_a^{l}(y_r^l))]+\mathbb E[log(1-D_a^{r}(y_r))],
\end{aligned}
\end{equation}
\begin{equation}\label{advD}
\begin{aligned}
\mathcal{L}_{adv}^{styD}=
-&\mathbb E[logD_b^{l}(x^l)]- {\mathbb E}[logD_b^{r}(x)]\\
-&\mathbb E[logD_a^{l}(y^l)]- {\mathbb E}[logD_a^{r}(y)]\\
-&\mathbb E[log(1-D_b^{l}(x_r^l))]
-\mathbb E[log(1-D_b^{r}(x_r))]\\
-&\mathbb E[log(1-D_a^{l}(y_r^l))]
-\mathbb E[log(1-D_a^{r}(y_r))],
\end{aligned}
\end{equation}
where low resolution auxiliary rendered photo $x_r^l=Dec_a^{l}(Enc_a(c_a\sim q_a(c_a|x)+s\sim N(0,I)))$, rendered photo $x_r=Dec_a^{r}(Enc_a(c_a\sim q_a(c_a|x)+s\sim N(0,I)))$,  {$\mathbb E$ means expectation.}
%
%
$y_r^l$, $y_r$, $y^l$ and $y$ in the caricature domain can be obtained in the same manner.

To preserve the identity information, inspired by Gatys \textit{et al.}~\cite{gatys2015neural}, we define the content loss to enforce the content similarity between the photo $x$ and rendered photo $x_r$,
\begin{equation}\label{content_loss}
\mathcal{L}_{cont}=\parallel \xi (x)-\xi (x_r)\parallel+\parallel \xi (y)-\xi (y_r)\parallel,
\end{equation}
where $\xi(\cdot)$ is a pretrained VGG net~\cite{simonyan2014very}.

\noindent
{\bf Contrastive Style Loss} 
%
%
To enforce the similarity of the rendered photo to caricatures and its discrepancy to the photos, we propose a constrastive style loss function to pull the texture of rendered photo $x_r$ to the input caricature and push it away from the input photo.
First, given two feature maps $m$ and $n$, the style distance~\cite{gatys2015neural} can be defined as,
\begin{equation}
\begin{aligned}
d(m,n)=\frac{1}{4*n_c*n_h*n_w}\sum\limits_{1}^{n_c}(G_{ij}^m-G_{ij}^n)^2,
\end{aligned}
\end{equation}
where $G^m$ is the gram matrix, which represents the dot product of the of feature $m$ in any two channels. $G^n$ is the gram matrix of feature $n$ extracted by a pretrained VGG net. 
%
Gram matrix measures the characteristics of each feature dimension and their relationships. Therefore, it can enlarge the texture details.
Then we define our contrastive style function as,
\begin{equation}
\begin{aligned}
Ctr(i_1,i_2,l)=&\frac{1}{2}[l\cdot d(i_1,i_2)^2+\\
&(1-l)max(mg-d(i_2,i_1),0)^2],
\end{aligned}
\end{equation}
where $l\in \{0,1\}$ is the label of the images pair \{$i_1,i_2$\}, the style distance between the image pair decreases when $l=1$, and increases when $l=0$. 
%
%
$mg>0$ is a margin, indicating that only the images from two domains with a style distance between 0 and $mg$ are considered. The whole contrastive style loss is,
\begin{equation}\label{ctr_loss}
\begin{aligned}
\mathcal{L}_{ctr} = &\alpha_1 Ctr(x_r,x,0)+\alpha_2 Ctr(x_r,y,1)+\\
&\alpha_3 Ctr(y_r,y,0)+\alpha_4 Ctr(y_r,x,1).
\end{aligned}
\end{equation}
where $\alpha_i$ is the hyper parameters.

The contrastive style loss in Eq.~\eqref{ctr_loss} pulls the style distance between rendered photo $x_r$ and input caricature $y$, while pushing its distance away from the input photo $x$ apart, which can lead to more plausible caricature style texture of rendered photo.
The rendered caricature $y_r$ is optimized in the same manner.

\subsection{Distortion Prediction Module}\label{DPM}
Despite of the plausible texture rendering, the other key issue in photo-to-caricature translation is the exaggerating deformation.
To obtain exaggerating deformation in a completely unsupervised fashion without any guidance, we propose a Distortion Prediction Module (DPM) is used in this stage. 
DPM accepts diverse distortion via a random-perturbed input photo.

DPM is a subnetwork with four fully connected layers. 
%
$DPM_a$ accepts photo as input, to output the corresponding displacements vector $v_a=\{v_a^1,v_a^2,\dots,v_a^n\}$.
we predefine the controlling points as $p_0=\{p^1,p^2,\dots,p^n\}$, where both displacement vector $v_a^i$ and controlling point $p^i$ are 2-D vectors in $u-v$ space. 
We employ the thin plate spline interpolation (TPS)~\cite{bookstein1989principal} for our deformation, due to its remarkable performance in warping~\cite{shi2019warpgan},
\begin{equation}
x^y=TPS_{p_0,v_a}(x_r).
\end{equation}

Now the rendered photo with exaggerating deformation is obtained. 
%
To make sure DPM can predict meaningful results, some constraints are needed to enforce decoder $Dec_b^{r}$ to generate plausible results. 
%
Discriminator $D_b$ is added in this stage to introduce the adversarial loss,
\begin{equation}
\begin{aligned}
\mathcal{L}_{adv}^{warpG}=\mathbb E[log(1-D_b(x^y))]+ \mathbb E[log(1-D_a(y^x))],
\end{aligned}
\end{equation}
\begin{equation}
\begin{aligned}
\mathcal{L}_{adv}^{warpD}=-&\mathbb{E}[D_b(x^y)]-\mathbb{E}[D_a(y^x)]\\
-&\mathbb E[log(1-D_b(x^y))]-\mathbb E[log(1-D_a(y^x))].
\end{aligned}
\end{equation}

Preventing from undesirable deformation, identity loss is introduced:
\begin{equation}\label{idtloss}
\mathcal{L}_{idt}=\mathbb{E}\parallel x^y-x\parallel + \mathbb{E}\parallel y^x-y\parallel.
\end{equation}

We set a small weight to this strong constraint to avoid generated effect from being too similar to the original image. 

\begin{figure*}[h]
	\centering
	\includegraphics[scale=0.42]{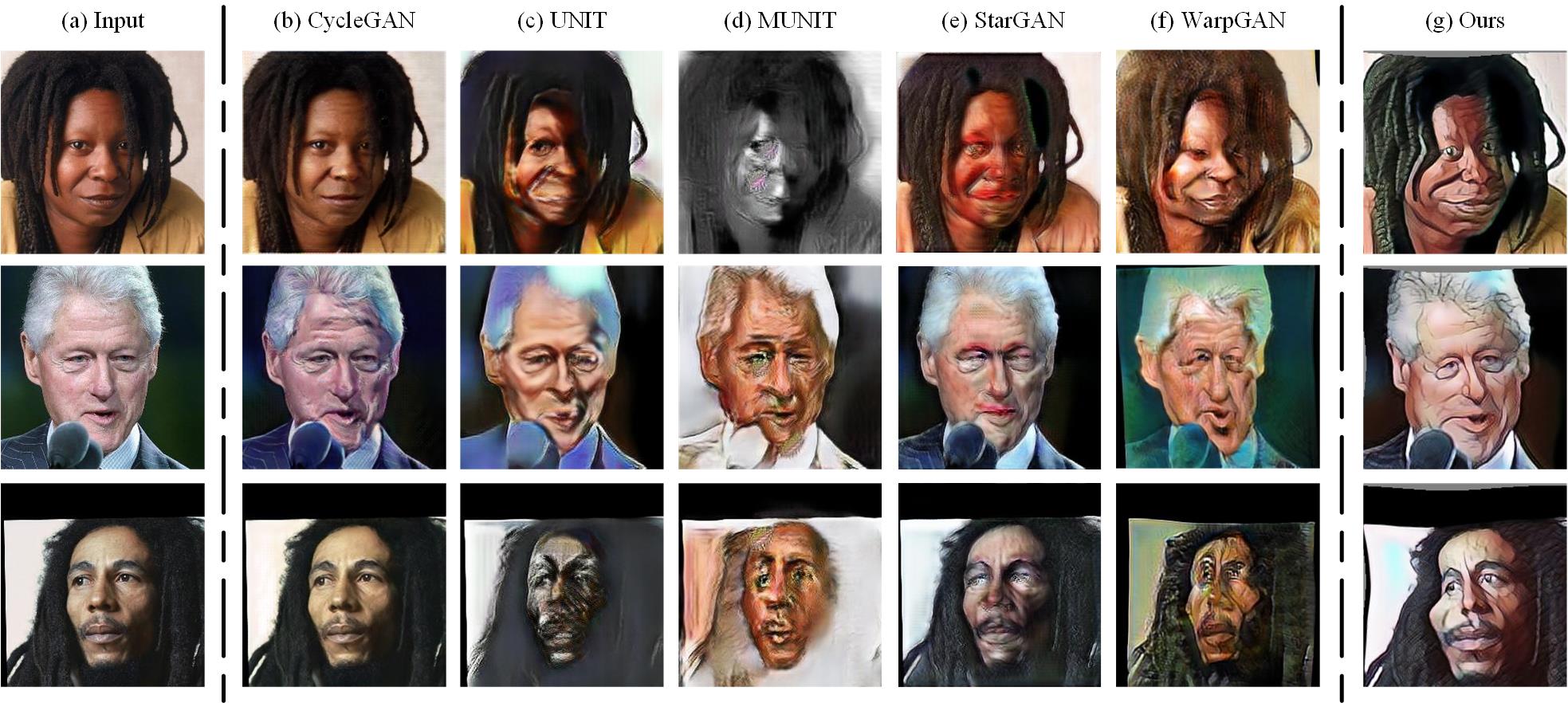}
	\captionof{figure}{Comparison with five state-of-art methods. (a) input photos, (b)-(f) Generation results of five state-of-art methods. (g) Our results.}\label{sota}
\end{figure*}

\subsection{Total Loss}
The proposed method is optimized in a two-stage strategy. First (section \ref{stage 1}), we optimize the style network via,
\begin{equation}
\begin{aligned}
\min\limits_{\tiny Enc_a,Enc_b,Dec_a,Dec_b}=&\lambda_r \mathcal{L}_{rec}+\lambda_K \mathcal{L}_{KL}+\lambda_a \mathcal{L}_{adv}^{styG}+\\
&\lambda_{c} \mathcal{L}_{cont}+\lambda_{ctr} \mathcal{L}_{ctr}, \\
\min\limits_{\tiny D_a^{l},D_a^{r},D_b^{l},D_b^{r}}=&\lambda_a \mathcal{L}_{adv}^{styD}.\label{eq15}
\end{aligned}
\end{equation}

Second, we optimize the distortion prediction modules via,
\begin{equation}
\begin{aligned}
\min\limits_{\tiny DPM_a,DPM_b}=&\lambda_a \mathcal{L}_{adv}^{warpG}+\lambda_i \mathcal{L}_{idt},\\
\min\limits_{\tiny D_a,D_b}=&\lambda_a \mathcal{L}_{adv}^{warpD}.\label{eq16}
\end{aligned}
\end{equation}

Note that we are under unsupervised setting, and support bidirectional translation.

\section{Experiments}\label{experiments}
We evaluate our method on the benchmark photo-caricature dataset Webcaricature~\cite{huo2017webcaricature}\cite{Huo:2017:VRC:3126686.3126736} consisting of 6042 caricatures and 5974 photos from 252 identities.

%
%
\subsection{Training Details}\label{trainingdetails}
 We employ Adam algorithm~\cite{kingma2014adam} to iteratively update our model according to the defined the loss functions Eq.~\eqref{eq15} and ~\eqref{eq16},  until achieving a minimal expectation. 
Hyper parameters $\beta_1$ and $\beta_2$ representing exponential decay rates is fixed as $\beta _1=0.5$ and $\beta_2=0.999$.

%
There are a random pair of photo and caricature in each mini-batch. 
%
We train our model for 100,000 and then 50,000  steps for the first and second stages respectively.
%
Learning rate is 0.0001.
 We first optimize our encoders, decoders, and discriminators except $D_a$ and $D_b$ via Eq.~\eqref{eq15},  then optimize DPMs, $D_a$ and $D_b$ via Eq.~\eqref{eq16} with the same learning rate.
We empirically set the hyper parameters \{$\alpha_1$, $\alpha_2$, $\alpha_3$, $\alpha_4$, $\lambda_r$, $\lambda_K$, $\lambda_a$, $\lambda_c$, $\lambda_{ctr}$, $\lambda_i$, $mg$\} as \{$0.5, 0.5, 1, 1, 10, 1, 1, 1, 0.5, 8, 2.0$\}.
The implementation platform is pytorch 1.4.0, python 3.6 on one Geforce GTX 1080 Ti GPU.

\subsection{Comparisons with State-of-the-Arts} 
We first qualitatively compare our caricature generation
method with four state-of-the-art methods as shown in Fig.~\ref{sota}.
We can see that, CycleGAN~\cite{zhu2017unpaired} and StarGAN~\cite{choi2018stargan}  tend to produce very photo-like images with trivial texture changes. This may be due to that these two methods are based on cycle consistency and multi-classification so that they retain too much photo’s information.
UNIT~\cite{liu2017unsupervised} and MUNIT~\cite{huang2018multimodal} demonstrate the most visually appealing texture styles due to the weigh-sharing strategy and AdaIN~\cite{huang2017arbitrary}, respectively. 
However, UNIT~\cite{liu2017unsupervised} generates too much artifacts without any consideration on geometric deformation.
AdaIN~\cite{huang2017arbitrary} assumes feature maps in different channels are uncorrelated, which ignores the global information and results in artifacts too.
All these four methods focus only on transferring the texture styles, while fail to deform the faces into caricatures. Although they try to compensate the distortion by using texture, it easily results in difficult training and mode collapse.
WarpGAN~\cite{shi2019warpgan} aims to generate plausible caricature, and take both texture and deformation into consideration. However, it also generates many unexpected artifacts as MUNIT~\cite{huang2018multimodal}. Since these two are both based on AdaIN~\cite{huang2017arbitrary}.
%
%
%
%
Our results in Fig.~\ref{sota} (g) present more plausible style and deformation with less artifacts, benefit from the texture rendering enforced by contrastive style loss on the weight-sharing strategy. In addition, the distortion prediction module allows each input photo an exaggerating deformation.

\subsection{Distortion Diversity}
\begin{figure}
\renewcommand\arraystretch{1.3}
\renewcommand\tabcolsep{1pt}
\center
\begin{tabular}{cccc}
	\hline
	(a) input &(b) $\alpha=10$ &(c) $\alpha=20$&(d) $\alpha=30$\\
	\includegraphics[scale=0.38]{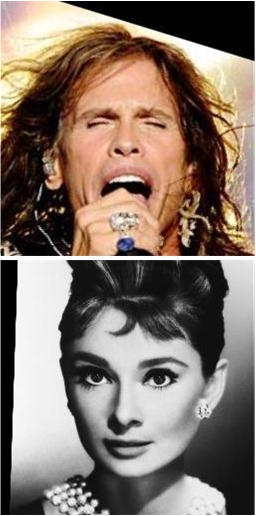}&
	\includegraphics[scale=0.38]{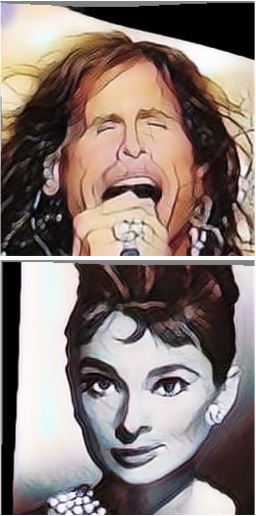}&
	\includegraphics[scale=0.38]{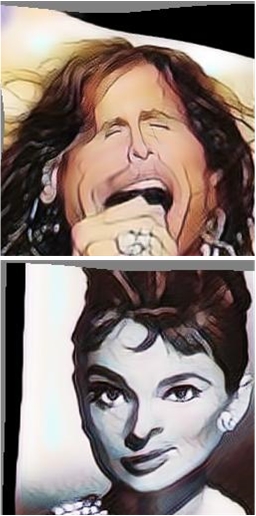}&
	\includegraphics[scale=0.38]{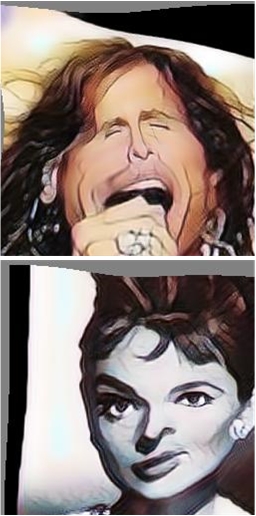}\\
	\hline
\end{tabular}
\caption{Examples of diverse distortions against the scale factor $\alpha$.}\label{factor}
\end{figure}

\begin{figure}
	\renewcommand\arraystretch{1.3}
	\renewcommand\tabcolsep{1pt}
	\center
	\begin{tabular}{cc}
		\hline
		(a) input & (b) with different gaussian noises \\
		\includegraphics[scale=0.38]{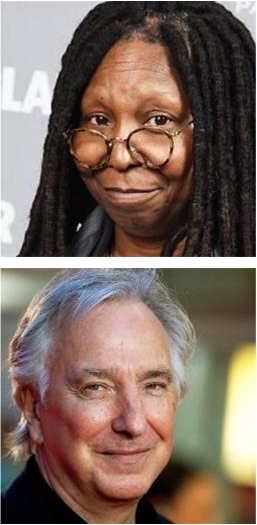}&
		\includegraphics[scale=0.38]{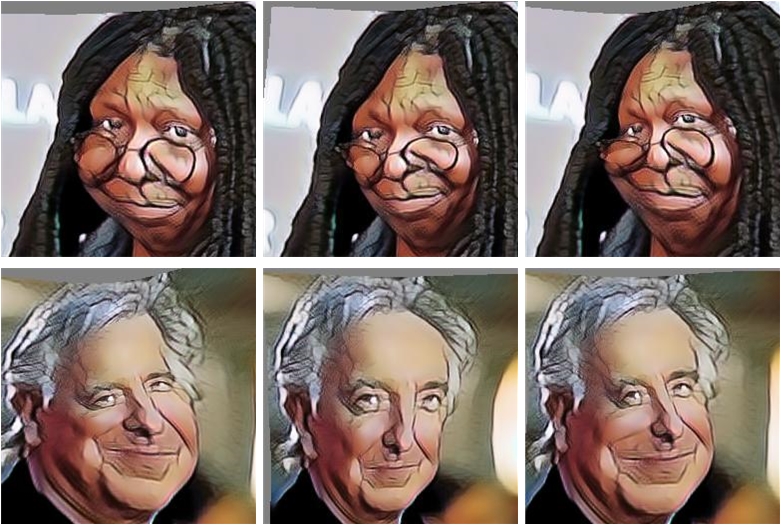}\\
		\hline
	\end{tabular}
	\caption{Examples of diverse distortions against different additional Gaussian noises.}\label{factor2}
\end{figure}

We introduce a scale factor $\alpha$ during deployment to allow customization of the exaggeration extent. We scale the displacement of control points i.e. warp vector $v$ by $\alpha $ to control how much the face shape will be exaggerated.
The bigger $\alpha$, the larger degree of deformation. 
Fig.~\ref{factor} shows the generated caricatures with different scale factors.
It is clear that the deformation increases as the scale factor ascends.
In addition, we sample different noises from standard Gaussian distribution with clamping in $[-0.1,0.1]$, and simply add it to input. DPM accepts a random-perturbed version of inputs, which cannot change the overall information but lead to more diverse deformation.
%
Fig.~\ref{factor2} demonstrates the diverse exaggeration results with additional random Gaussian noises.

\subsection{Ablation Study}\label{ablation study}
\begin{figure}
\renewcommand\arraystretch{1.3}
\renewcommand\tabcolsep{1pt}
\begin{tabular}{cccc}
\hline
(a) input &(b) w/o $\mathcal{L}_{cont}$ &(c) w/o $\mathcal{L}_{ctr}$&(e) $\mathcal{L}_{cont}+\mathcal{L}_{ctr}$ \\
\includegraphics[scale=0.38]{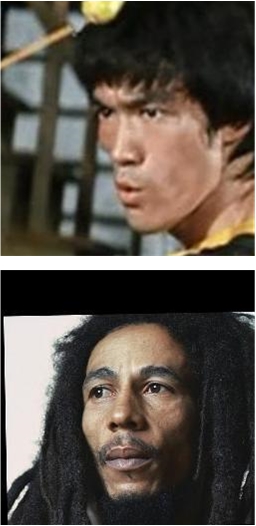}&
\includegraphics[scale=0.38]{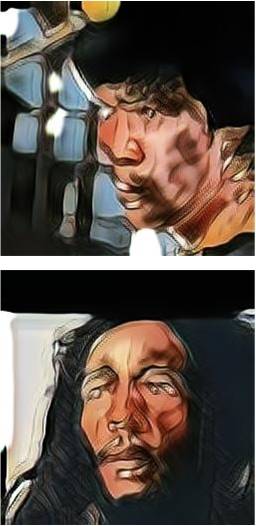}&
\includegraphics[scale=0.38]{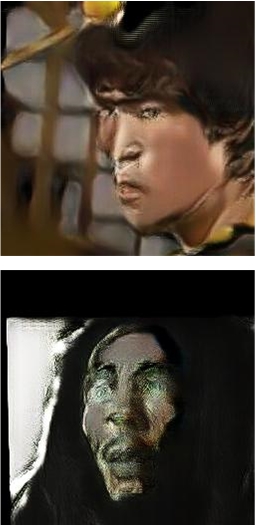}&
\includegraphics[scale=0.38]{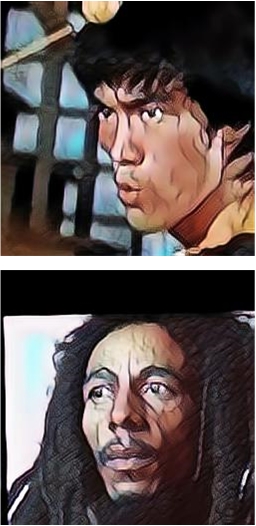}\\
\hline
\end{tabular}
\captionof{figure}{Ablation study on contrastive style loss $\mathcal{L}_{ctr}$ and content loss $\mathcal{L}_{cont}$. Note that all the results are the rendered images in the first stage.}\label{ablation}
\end{figure}

\begin{figure}
	\renewcommand\arraystretch{1.3}
	\renewcommand\tabcolsep{1pt}
	\center
	\begin{tabular}{cccc}
		\hline
		(a) input &(b) w/o $\mathcal{L}_{idt}$ &(c) Ours\\
		\includegraphics[scale=0.38]{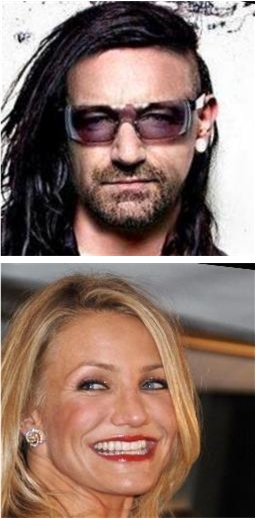}&
		\includegraphics[scale=0.38]{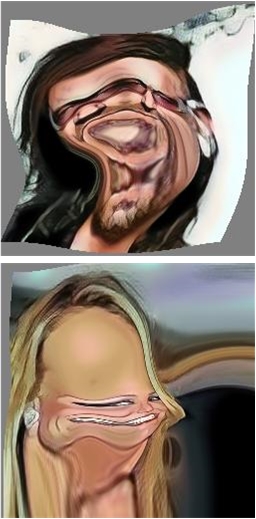}&
		\includegraphics[scale=0.38]{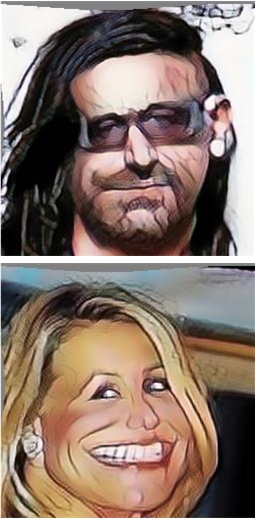}\\
		\hline
	\end{tabular}
	\captionof{figure}{Ablation study on identity loss $\mathcal{L}_{idt}$.}\label{ablation2}
\end{figure}

To verify the contribution of each component in our model, we evaluate three variants of the proposed model, by removing $\mathcal{L}_{ctr}$ and $\mathcal{L}_{cont}$ from the first stage, and removing $\mathcal{L}_{idt}$ from the second stage respectively for ablation study.
Fig.~\ref{ablation} and Fig.~\ref{ablation2} demonstrate several qualitative results of each variant.
Note that we only use the contrastive style loss $\mathcal{L}_{ctr}$ and the content loss $\mathcal{L}_{cont}$ in the first stage, therefore, as shown in Fig.~\ref{ablation}, we evaluate their contributions through the rendered images without deformation for better comparison.
It is clear that, Fig.~\ref{ablation} (b) indicates that model with content loss $\mathcal{L}_{cont}$ preserves more identity details between the generated images and the original ones, since their deep features are pulled together by $\mathcal{L}_{cont}$.
Fig.~\ref{ablation} implies that the contrastive style loss $\mathcal{L}_{ctr}$ plays a important role in image stylization, while it will result in poor style generation without $\mathcal{L}_{ctr}$
Fig.~\ref{ablation2} illustrates the results of the variant without identity loss $\mathcal{L}_{idt}$, from which we can see, it can lead to less unexpected deformation.



\begin{table}
	\renewcommand\arraystretch{1.3}
	\center
	\setlength{\tabcolsep}{14mm}{
		\begin{tabular}{c|c}
			\hline
			Method &Proportion \\
			\hline
			UNIT\cite{liu2017unsupervised}&24.39\%\\
			WarpGAN\cite{shi2019warpgan}&24.56\%\\
			Ours&51.05\%\\
			\hline
		\end{tabular}
	}
	\captionof{table}{Voting results of our user study.}\label{user}
\end{table}

\subsection{User Study}
We investigate the user study against two competitors,  WarpGAN~\cite{shi2019warpgan} and UNIT~\cite{liu2017unsupervised}.
42 caricatures synthesized by each of the three methods are presented to 29 subjects, who are told to vote the best caricature considering both geometry and style aspects.
The results are shown in Table~\ref{user} reports the voting ratios of each method.
Our method gets the highest voting against the other two competitors, which verifies the performance of our method on photo-caricature generation.

\subsection{Face Recognition}\label{faceRecognition}
To evaluate the ability of identity maintaining of our method, we evaluate the face recognition task on the whole dataset via the widely used face recognition model SphereFace~\cite{liu2017sphereface}. 
Specifically, we select one photo for each 252 identity in training and testing sets as gallery, while randomly selecting 4000 photos, hand-drawn caricatures, caricatures synthesized by WarpGAN and our methods respectively as probe. 
Table~\ref{faceRec} reports the Rank-1 accuracy of the recognition results in four probe scenarios.
It is clear that, caricatures are harder to preserve the identity information comparing to the photos due to the large distortions and huge style changes. 
Our method still achieves comparable accuracy as WarpGAN and significantly beats results probed by the hand-drawings, which verifies the ability of identify preservation of our method.

\begin{table}
	\renewcommand\arraystretch{1.3}
	\center
	\setlength{\tabcolsep}{10mm}{
		\begin{tabular}{c|c}
			\hline
			Probe &Rank-1 accuracy \\
			\hline
			Photos & 100\%\\
			Hand-drawnings &8.46\%\\
			WarpGAN~\cite{shi2019warpgan} &34.18\%\\
			Ours  &34.56\%\\
			\hline
		\end{tabular}
	}
	\captionof{table}{Rank-1 face recognition accuracy of four different matching protocols using a state-of-art face recognition model SphereFace~\cite{liu2017sphereface}.}\label{faceRec}
\end{table}

\subsection{Caricature-to-Photo Translation}
\begin{figure}
	\centering
	\includegraphics[scale=0.36]{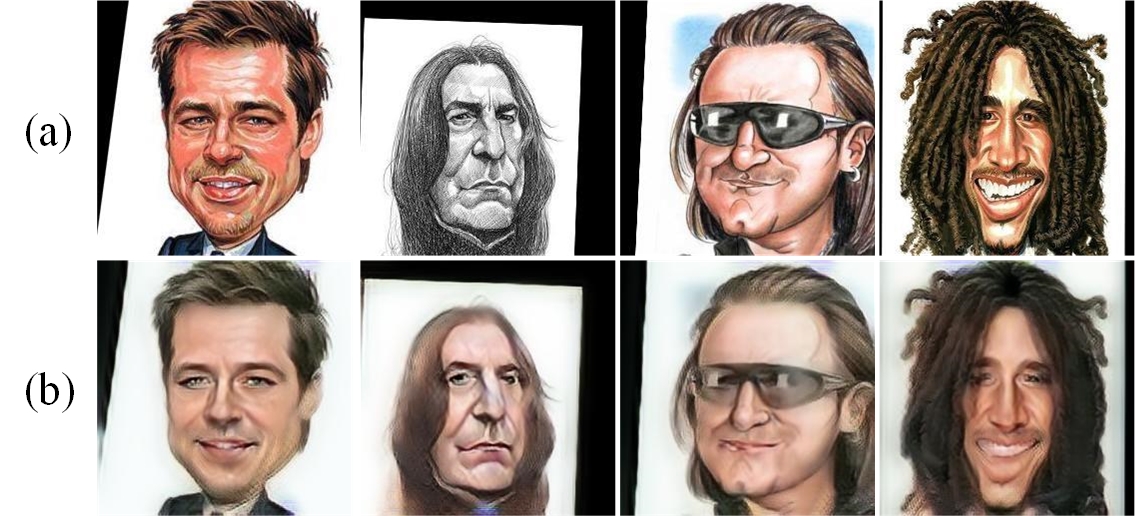}
	\captionof{figure}{Caricature-to-photos synthesizing without deformation. (a) Input caricatures. (b) Transformed photos.}\label{b2a}
\end{figure}

Benefit from our symmetric structure, our method supports the bidirectional translation between photos and caricatures.
Fig.~\ref{b2a} demonstrates additional results of caricature-to-photo synthesizing. However, our distortion modules are not well suitable for the caricature-to-photo translation since caricature's shape is extremely irregular.
Results show that the proposed method can learn the skin texture with fewer artifacts.
\section{Conclusion}
In this paper, we propose a symmetric architecture for photo-to-caricature translation. We divide this process into two stages, style rendering and geometry distortion. In the first stage, we propose a novel contrastive style loss to better render the texture style and decrease the artifacts. In the second stage, we propose a distortion prediction module for diverse unsupervised deformation. Comprehensive experiments verify the effectiveness of our method comparing with the existing methods.

\section{Acknowledgements}
This work is partially funded by Beijing Natural Science Foundation (Grant No. JQ18017), Youth Innovation Promotion Association CAS (Grant No. Y201929), the National Natural
Science Foundation of China (61976002) and the Natural Science Foundation of Anhui Higher Education Institutions of China (KJ2019A0033).







%
%
%

\bibliographystyle{IEEEtran}
\bibliography{IEEEabrv,reference}

\clearpage

\end{document}